# Optimal Planning and Machine Learning for Responsive Tracking and Enhanced Forecasting of Wildfires using a Spacecraft Constellation

Sreeja Roy-Singh[1,2], Vinay Ravindra[1,2], Richard Levinson[1,3], Mahta Moghaddam[4], Jan Mandel[5], Adam Kochanski[6], Angel Farguell Caus[6], Kurtis Nelson[7], Samira Alkaee Taleghan[5], Archana Kannan[4], Amer Melebari[4]

[1] NASA Ames Research Center, Moffet Field, CA 94036, USA
[2] Bay Area Environmental Research Institute, Moffet Field, CA 94036, USA
[3] KBR Wyle Services, 601 Jefferson Street, Houston, TX 77002, USA
[4] University of Southern California, 3470 Trousdale Parkway Los Angeles, CA 90089, USA
[5] University of Colorado, 2055 Regent Drive, Boulder, CO 80309, USA
[6] San Jose State University, 1 Washington Square, CA 95192, USA
[7] United States Geological Survey, Sioux Falls, SD 57198, USA
`sreeja.nag@nasa.gov`

**Abstract.** We propose a novel concept of operations using optimal planning methods and machine learning (ML) to collect spaceborne data that is unprecedented for monitoring wildfires, process it to create new or enhanced products in the context of wildfire danger or spread monitoring, and assimilate them to improve existing, wildfire decision support tools delivered to firefighters within latency appropriate for time-critical applications. The concept is studied with respect to NASA's CYGNSS Mission – a constellation of passive microwave receivers that measure specular GNSS-R reflections despite clouds and smoke. Our planner uses a Mixed Integer Program formulation to schedule joint observation data collection and downlink for all satellites. Optimal solutions are found quickly that collect 98-100% of available observation opportunities. ML-based fire predictions that drive the planner objective are >40% more correlated with ground truth than existing state-of-art. The presented case study on the TX Smokehouse Creek fire in 2024 and LA fires in 2025 represents the first high-resolution data collected by CYGNSS of active fires. Creation of Burnt Area Maps (BAM) using ML applied to the data during active fires and BAM assimilation into NASA's Weather Research and Forecasting Model using neural networks to broadcast fire spread are novel outcomes. BAM and CYGNSS-obtained soil moisture are integrated for the first time into USGS fire danger maps. Inclusion of CYGNSS data in ML-based burn predictions boosts accuracy by 13%, and inclusion of high-resolution data boosts ML recall by another 15%. The proposed workflow has an expected latency of 6-30h, improving on the current delivery time of multiple days. All components in the proposed concept are shown to be computationally scalable and globally generalizable, with sustainability considerations such as edge efficiency and low latency on small devices.

**Keywords:** Artificial intelligence, satellite remote sensing, wildfire predictions.

# 1 Introduction

Wildfires are a serious problem for humans and wildlife, and the impact in the past few decades has grown in terms of area affected and cost to suppress [1]. More than a third of our housing units lie on the wildland-urban interface, thereby increasing the risk of ignition and urban spread [2]. In 2024 alone [3], there were 64,897 wildfires that burned 8.9 million acres in the U.S. Satellite remote sensing is invaluable for tracking the spread of large wildfires owing to its global coverage from a high vantage point and regular revisit of the same ground location. Constellations of low-cost satellites in Low Earth Orbit (LEO) offer unprecedented temporal revisits at 10x closer distances than traditional weather satellites in geostationary orbit. When equipped with passive microwave sensors, they can measure through smoke, clouds, and canopy. The sensor observations, when optimally collected using a novel, proposed AI-based planner and processed using novel, proposed ML-based models, yield fire danger and spread metrics that are functionally novel and/or higher quality. We demonstrate the benefit of the proposed AI/ML technologies by applying the full workflow to pre-fire conditions and active fires over 3 days in the contiguous United States (CONUS). The benefits of active fire ML products are further demonstrated on spacecraft-collected data over the Smoke House Creek fire, which burned more than a million acres in 2024 and was the largest wildfire in Texas history [4], and the Los Angeles fires in 2025 which caused damages over $250 billion, one of the costliest disasters in U.S. history [5]. These geographies (CONUS or 2 wildfires) or time periods (3 days) are examples to demonstrate the value of the technologies, which can be applied to any region, number of fires, or time-period.

## 1.1 Technology Challenges

The challenges associated with time-critical collection and actionable utilization of spaceborne observations are on both the space technology and the ground logistics sides. Artificial Intelligence (AI) can be applied at multiple stages of the data pipeline from space to actions on Earth and we demonstrate the value in doing so at every stage using novel technologies.

1. *Data Collection:* Appropriate satellite observations are constrained by hardware resources (energy, data) and orbital dynamics (access to observation opportunities and uplink/downlink with Earth) [6]. LEO satellites provide limited Earth coverage at any time. Automated planners can prioritize observations of specific locations and times, given constraints such as limited onboard storage capacity. Collection of high-resolution imagery (e.g., rawIF data, see Section 2) is therefore resource-limited and requires multiple passes over ground stations to be downlinked to Earth. AI can be used in scheduling to improve collection and downlink quality and efficiency.

2. *Data Processing:* Remote sensing data can be captured at frequencies beyond human vision, and sensors detect intricate phenomena that are not discernible to the human eye. Machine Learning (ML) is a powerful tool for extracting meaningful information from complex sensory data, such as Delay-Doppler maps (DDMs, as seen in Fig. 1 inset) from passive microwave signals [7], and creating useful geospatial products such as Burnt Area Maps (BAMs) and Soil Moisture (SM) maps from DDMs .

3. *Data Application to forecasts:* High performance computing (HPC) is necessary for assimilating ancillary, real-time parameters into forecast models such as the WRFx forecasting system, an open-source framework that uses WRF-SFIRE, a high resolution, multi-scale weather forecasting model, coupled with a semi-empirical fire-spread model and a prognostic dead fuel moisture model [8]. ML supports improved predictions of active fire pixels and Fire Arrival Time (FAT) by fusing data from various sensor sources within WRFx and using a multi-head neural network toward classification and regression. This supports firefighters and may provide feedback for the next iteration of intelligent planning to gather responsive data.

## 1.2 Multidisciplinary, Multilateral Approach

Our goal is to monitor from space, areas that are known to be experiencing severe active fire activity as well as regions at risk of imminent large wildfires. For the proposed concept of operations (ConOps) to provide benefit for on-ground firefighters requires collaborative effort from researchers of several disciplines and institutions. Our collaboration for the presented case study includes:

- Computer scientists for spacecraft planning and scheduling in NASA Ames Research Center (ARC)
- Instrument scientists from NASA Jet Propulsion Laboratory and universities
- Astrodynamics and spacecraft designers from the Bay Area Environmental Research Institute in NASA ARC
- Atmospheric Scientists from the Wildfire Interdisciplinary Research Center, San Jose State University
- HPC Computer scientists in universities for integration into operational wildfire models
- Physical scientists and Incident Management Team (IMT) support from U.S. Geological Survey (USGS)

IMT and USGS also use our data products on ground and can provide actionable feedback to improve the technologies.

## 2 Concept of Operations

As illustrated in Fig. 1, a distributed spacecraft network can make observations that are periodically downlinked to ground stations (see grey box on "Operations"). All computation to plan those observations and process their data (outside the grey box) runs offboard on Earth and resultant plans are uplinked to the spacecraft when it accesses ground stations. The Planner generates a sequence of feasible satellite observations and downlink operations. The Planner is guided by (A) location-based priorities quantified by the "Priority Framework.", and (B) constraints imposed by physics and hardware limitations, quantified by the "Satellite Orbital Simulator" which predicts observation and downlink opportunities. Collected data is first converted to geophysical products in the "Data Processing" block, then integrated into fire models (pre-fire danger in orange and active-fire in red) to improve wildfire forecasts. These enhanced forecasts support stakeholders in making critical fire management decisions and input to the Planner for the next horizon of scheduling. This feedback loop enables the planner to assess gaps, re-forecasts, and intelligently plan the next iteration. More details on the concept are found in Ref [9].

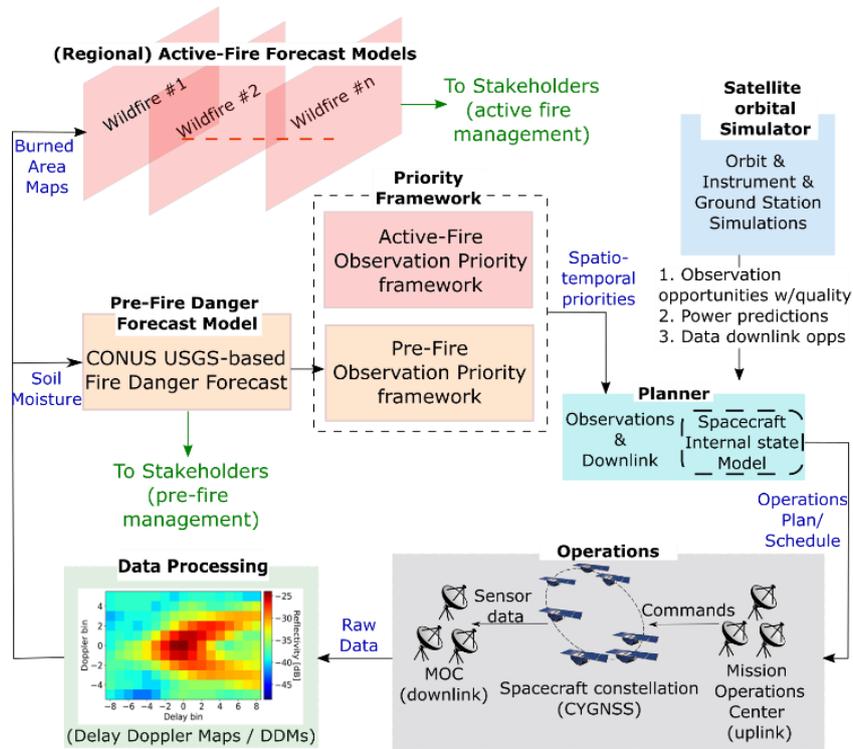

**Fig. 1.** System Architecture and Concept of Operations (ConOps).

We apply the ConOps to a constellation of 8 designed satellites (of which 7 are active) carrying Global Navigation Satellite Systems Reflectometry (GNSS-R) sensors. GNSS-R measures properties of surface specular points using reflected GNSS signals. It has demonstrated enormous scientific value via NASA's Cyclone Global Navigation Satellite System (CYGNSS) [7] and commercialized by companies like Spire and Muon. The low size, weight, power (SWaP) needs for GNSS-R sensors allows them to be accommodated on small satellites, which enables deployment of numerous copies and enables frequent sampling (observations) globally. While having a small SWaP reduces the overall cost (development, manufacturing, launch) of the constellation, it constrains onboard resources.

For example, CYGNSS usually gathers data in a 'nominal' mode - the raw observation signals are processed onboard and downlinked with information loss but substantially saving onboard data storage and downlink. It is possible to command a 'RawIF' mode of observation, in which the raw signals are downlinked for more informative retrievals on ground, albeit at the cost of 1680x more storage and downlink time required. Only 60sec worth of rawIF images can be stored onboard, which takes ~20min to downlink spread over several hours of access gaps. We seek to command RawIF signals over locations of higher wildfire priority within tight resource and access constraints, therefore the necessity of an AI-planner. The novel geophysical products created after ML-based processing are BAM for active fires and SM of pre-fire unburnt areas. BAM and SM are ingested by a neural network, that simultaneously detects active fires and predicts their arrival times at improved accuracy, and a pre-fire danger predictor, respectively.

Our case study focuses on three dates of significant fire activity over the United States with listed peak fire locations:
A. 8 Jan 2025: Los Angeles (LA) fires, comprised of the Palisades, Eaton and Hughes fires in California (CA)
B. 26 Feb 2024: Smokehouse Creek fire in Texas (TX)
C. 5 July 2023: Day with the highest number of wildfires in CONUS at latitudes less than 35 deg (relevant because CYGNSS orbits are low inclination)

Sections 3-5 describe the three parts of the ConOps, components within them, and the novel AI elements being developed. We describe results from each as would be executed sequentially. The feedback cycle in Fig. 1 is expected to execute every 24h and all planning / processing to take a few hours as validated in our results. Quicker cycles are possible at the cost of more computation, space-to-ground access contacts, and execution resources.

While similar ConOps have been demonstrated to monitor volcanos [10] and urban floods [11], we present a novel application as enabled by the novel technology described in Section 3-5. This represents the ***first*** ever collection of spaceborne GNSS-R raw-IF data for unprecedented passive-microwave resolution of active wildfires, and the ***first*** ML-based generation of medium-spatial, high-temporal resolution BAMs as the fire progresses. Our Mixed Integer Linear Programming (MILP) planner formulation and plan represents the ***first*** optimal solution for a constellation of active, agile satellites where observation and downlink are optimized simultaneously using real-world data-driven prediction models. This ConOps also demonstrates the ***first*** assimilation of any GNSS-R dependent product into ***both*** pre-fire and active fire decision tools: [i] USGS danger maps released daily over CONUS, and [ii] WRFx and ML-based active fire forecasts on fire area extent and progression (outputs are available publicly via Ref [12]).

## 3    Observation Prioritization Framework

Appropriate quantification of observational priorities is the backbone for *gathering spaceborne data efficiently* for either active fires or pre-fire risk conditions. AI-driven improvements in priorities and uncertainties significantly impacts the quality and responsiveness of subsequent observations.

### 3.1    Active Fire Prioritization (WRFx fire danger)

The state of art (Fig. 2 top half) for evaluating active wildfire danger is non-standardized and varies between companies, applications, users. One of the most advanced products, although experimental, is WRFx and its Fuel Moisture Data Assimilation (FMDA) system, which updates hourly and forecasts *dead fuel* moisture for the next 48 hours. FDMA integrates measurements from ground-based Remote Automated Weather Stations (RAWS) and High-Resolution Rapid Refresh (HRRR) of weather data into a fuel moisture model. We propose extending this system by developing a new active fire danger quantification (purple box) that uses multiple Fire Weather Indices (FWIs) *in addition* to HRRR weather data and FMDA fuel moisture estimates. Using this new product and real-time wildfire ignition location data from the National Interagency Fire Center (red), active fire event priority (pink) will be published at hourly cadence.

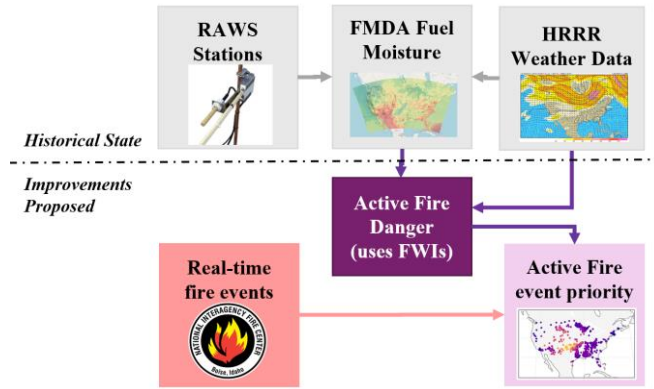

**Fig. 2.** Algorithmic workflow for Active Fire prioritization

An ML model utilizing XGBoost [13] is being developed for active fire event prioritization, which integrates several state-of-the-art FWIs listed in Fig. 3's X-axis. The model trains on historical wildfire activity data and learns the characteristics of wildfires that grow large and become difficult to control. Model validation shows accuracy at 0.86, recall at 0.87, precision at 0.73. While there is much scope to improve quality by extending to a more diverse set of FWIs and improving XGboost features, the model already outperforms all major FWIs used for fire danger predictions in terms of matching predictions with ground truth. Fig. 3 shows the number of large and small fires predicted by the model compared to many FWIs (KBDI, FFWI, LFP, Haines, DMC,

FFMC, ISI, SAWTI) and ground truth. The Matthew's Correlation Coefficient (MCC) of the ML model output with respect to ground truth at 0.75 is better than the MCC of all other FWIs. The MCC is >40% higher than the next highest MCC associated with an FWI. The model is also computationally efficient: 1-hour processing time for every set of 64 parallel active fires.

The model outputs CONUS location priorities for our case study (Fig. 4). The most destructive wildfires on that day were confirmed to be associated with higher priority values. For example, the 2025 Palisades fire recorded a value of 0.72, the 2025 Eaton fire at 0.77, the 2024 Smokehouse Creek fire at 0.87, and the 2023 Lincoln fire at 0.74 represent the highest danger. As model quality improves, we expect to quantify wildfire severity and behavior more robustly.

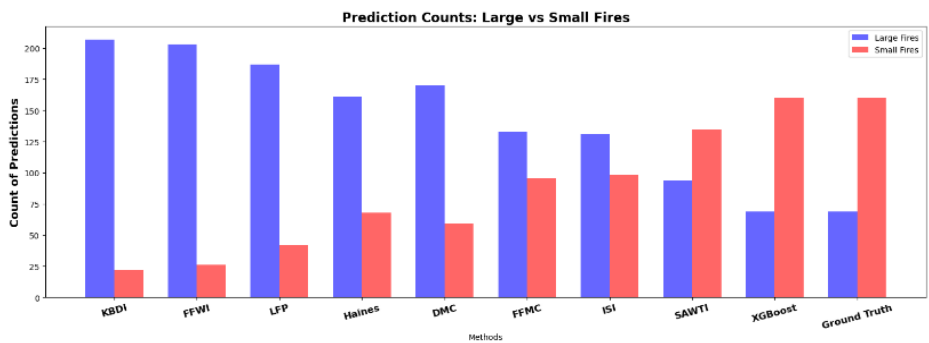

**Fig. 3.** Comparison of the number of large and small figures predicted by the WRFx Fire Danger ML model (labeled 'XGBoost') compared to 8 state-of-art FWIs and to ground truth.

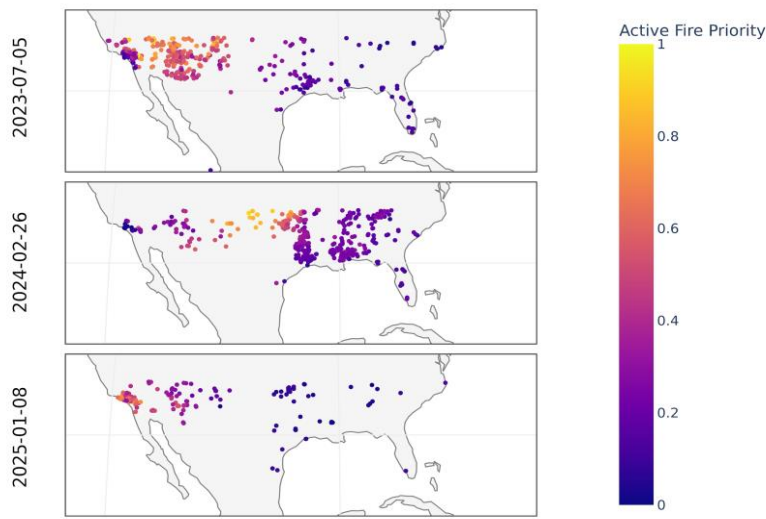

**Fig. 4.** Active Fire event priority across CONUS for three selected dates

### 3.2    Pre-Fire Prioritization (USGS Fire Danger)

The USGS produces daily 7-day forecasts of potential large fire activity for CONUS based on satellite-derived pre-fire products such as vegetation greenness, fuel moisture, and predicted weather conditions [14]. These data are used to produce the Wildland Fire Potential Index (WFPI), a unitless value that is correlated with increased incidence of large (>500 acres) wildland fires [15]. WFPI-derived Large Fire Probability (WLFP) and WFPI-derived Fire Spread Potential (WFSP) [16,17] are predictive indices derived from daily WFPI forecasts. WLFP is the chance that a large fire will occur in a specific location at a specific time based on historical fire activity patterns adjusted to account for the current WPFI values. WFSP is the chance that an existing wildland fire ignition will escape initial attack and grow to be large based on relationships between historical ignitions and WFPI. These indices are used to inform the pre-fire priority of any point in CONUS to optimally plan CYGNSS operations.

## 4      Planning Satellite Operations

An automated, intelligent planner is used to schedule observations over active fires and pre-fire conditions. Previous studies [11,20] have shown that even in the lack of optimality guarantees, intelligent planners significantly improve performance compared to manual, static scheduling.

The *planner's input* includes timepoints (seconds) when each satellite may observe geographic locations or downlink data to a ground station (GS). The open-source Earth Observation Simulator (EO-Sim) software is used to predict observation opportunities for target coordinates on Earth, downlinks to GS, and periods of eclipse which is used in the calculations of the battery state [18,19]. The orbit propagator provides satellite states at discrete time points in 1 sec interval over a pre-defined planning horizon (24h for the case study). EO-Sim calculates observation opportunities based on the computed specular (reflected) locations, i.e., reflection points from pairs of GNSS and CYGNSS satellites, for any selected geography.

*Spacecraft Storage and Energy State Models:* Each RawIF image spans 1 second of data collection and consumes 1/60 of data buffer capacity. The buffer is a First-In-First-Out queue and can store a maximum of 60 secs worth of RawIF data. It takes ~20s to downlink a single image, and ~20 min to download a full buffer. Power is produced continuously by solar panels, except during eclipse (~30 of ~90 min). Power is consumed continuously by nominal operations, and additionally for data collection and downlink. RawIF and nominal mode images are stored in different buffers.

The *planner's output* is a command schedule for each satellite, each second when it should collect or downlink data.

*Large Search Space:* The raw inputs identify commands at 1-sec granularity, related to the CYGNSS satellite's speed in LEO (~7 km/s). For each satellite, there are ~23,000 seconds when it may collect over CONUS or downlink data or remain idle. This sequence of ~23k binary choices per satellite creates a huge search space. Prior time-indexed approaches to this problem modeled each second scaled poorly or formulated the problem non-optimally [20].

*Data Cycle Abstraction*: We developed a novel interval-based abstraction which improves scaling and enables optimal solutions to be found quickly. Instead of decision variables indexed at the 1-sec granularity, we model temporal intervals called Data Cycles, constructed from the raw orbital access times during preprocessing. Each data cycle is a sequence of 2 phases: First the observation phase fills up storage, followed by the downlink phase which frees up storage. The observation time window is associated with a unique set of images available during that interval. An image can cover multiple targets or geo-locations. Rewards are associated with targets and there are many targets per image, while storage is consumed by images. Cycles are repeated over the plan horizon. In our case study, satellites have an average of 6.5 data cycles every 24 hours.

### 4.1    Planning Model for Observation & Downlink

The planner has been implemented as a MILP  formulation, generalized for any geography, spacecraft, payload, ground station, with agility options. When applied to our case study, optimal solutions are found within an hour. We highlight the key elements below due to lack of space to describe the full formulation. The planner's *objective* is to maximize the aggregate science rewards collected for all observed targets across all satellites. The planner chooses which satellites will observe which targets, and when to observe them. Each observation collects one *image*, which covers multiple *targets*. The same target may be covered by multiple images.

**Parameters:**
A = {set of active fire targets}
P = {set of pre-fire targets}
T = {set of all targets: $A \cup P$}, $m \in T$ is a target
I = {set of unique image IDs for each image opportunity}
$I_{s,k}$ = images visible to satellite $s$ during cycle $k$
$I_m$ = images which cover target $m \in T$
$r_a^A$ = reward for active fire target $a \in A$. This is the active fire target prioritization value described in section 3.1.
$r_b^P$ = reward for pre-fire target $b \in P$. This is the pre-fire prioritization value described in section 3.2.

**Binary Decision Variables:**
$x_m$ = 1 if target $m \in T$ is observed, otherwise = 0
$w_i$ = 1 if image $i \in I$ is collected, otherwise = 0

**Continuous Decision Variables:**
$sa_{s,k}$ = storage available on sat $s$ at the beginning of cycle $k$
$sc_{s,k}$ = storage consumed (used) on sat $s$ during cycle $k$

$sp_{s,k}$ = storage produced (freed up) on sat $s$ during cycle $k$
$ea_{s,k}$ = energy available on sat $s$ at the beginning of cycle
$eNet_{s,k}$ = net change in energy on sat $s$ during cycle $k$

**Objective:**

$$\text{Maximize: } (10 * \Sigma_{a \in A} r_a^A x_a) + \Sigma_{b \in P} r_b^P x_b$$

$$(1)$$

**Constraints:**

$$x_m \leq \Sigma_{i \in I_m} w_i \quad \forall m \in T$$

$$(2)$$

$$x_a = 1 \qquad \forall a \in A$$

$$(3)$$

Our goal is to schedule the active fire targets first and then use remaining resources for pre-fire observations. Equation (1) shows the objective is a weighted sum of the rewards for the active and pre-fire targets observed which favors active fire targets by a factor of 10. Equation (1) ensures target rewards are counted only once in the objective, even when observed by multiple satellites. This produces coordinated observations between satellites. Constraints (2) ensure that if target $m$ is observed, then at least one image $i$ covering target $m$ must be planned. Constraints (3) say all active fire targets must be scheduled. This hard constraint treats active fire targets as "mandatory", but the constraint itself is optional. If it is not possible to observe all targets, the solver will immediately return "*infeasible*", in which case we can remove this hard constraint and fall back on the weighted sum of active and pre-fire target values in the objective (1). In our case study, it was always possible to observe all active fire targets hence constraint (3) was always included.

The planner enforces these additional constraints:
1. No data collection allowed when storage is full
2. No downlinks allowed when storage is empty
3. Battery charge can never dip below a minimum level
4. Only one satellite at a time can downlink to each GS

Our novel *data cycle* abstraction makes the problem more tractable. The planner tracks aggregate resource (energy, storage) production and consumption during each cycle interval, without modeling individual seconds, as a function of time spent collecting and downlinking data. Resource constraints are enforced at the end of each data cycle.

### 4.2    Planner Evaluation

All experiments were run on a 2023 MacBook Pro, with an M3 Max chip, 36 GB Memory, using Python 3.8. The MILP solver was Gurobi version 11.0.3 [21]. MILP solvers may be stopped at any time and produce a feasible solution (hence responsive architecture), and solution quality monotonically improves with planning time.

Experiments using the planner and evaluating it are summarized in Table 1. For each date in the case study, we ran three tests. Active & Pre tests include both active fire and pre-fire targets, and use the Gurobi solver's default MIP gap tolerance, i.e. difference between the current best objective score and the solver's estimate of the optimal score, of 1e-4. The solver terminates with an optimal solution when the gap dips below the tolerance. The *Active and Pre* (0.01) tests relax the MIP gap tolerance to 1e-2 instead of 1e-4. The Active Only tests include only the Active Fire targets. Table 1 shows the MIP gap when the solver terminated, either when it timed out after 3-hours or when an optimal solution was found. Optimal cases are marked by (*) and colored text in blue.

| Test: | MIP Gap | Solve Time | Science Rewards | | | Active Fire Targets | | | Pre-Fire Targets | | |
|---|---|---|---|---|---|---|---|---|---|---|---|
| | | | Plan (objective) | Avail-able | Rwd % | Plan | Avail-able | Target % | Plan | Avail-able | Target % |
| **LA Fires, CA: 1/8/2025** | | | | | | | | | | | |
| **1**: Active & Pre | 0.005 | 3 h | 47993 | 48345 | 99.3 | 136 | 136 | 100 | 15209 | 19034 | 79.9 |
| **2**: Act & Pre (0.01)* | 0.0086 | 41 m | 47863 | 48345 | 99.0 | 136 | 136 | 100 | 14434 | 19034 | 75.8 |
| **3**: Active Only * | 0.0001 | 0.01 s | 46741 | 46741 | 100 | 136 | 136 | 100 | N/A | N/A | N/A |
| **Smokehouse, TX: 2/26/2024** | | | | | | | | | | | |
| **4**: Active & Pre | 0.007 | 3 h | 156294 | 157968 | 98.9 | 359 | 359 | 100 | 19702 | 22119 | 89.1 |
| **5**: Act & Pre (0.01)* | 0.0099 | 22.6 m | 155856 | 157968 | 98.7 | 359 | 359 | 100 | 19427 | 22119 | 87.8 |
| **6**: Active Only * | 0.0001 | 0.01 s | 108046 | 108046 | 100 | 359 | 359 | 100 | N/A | N/A | N/A |
| **Most fires in 1 day: 7/5/2023** | | | | | | | | | | | |
| **7**: Active & Pre | 0.0023 | 3 h | 208727 | 209617 | 99.6 | 367 | 367 | 100 | 19871 | 23216 | 85.6 |
| **8**: Act & Pre (0.01)* | 0.0056 | 22.9 m | 208089 | 209617 | 99.3 | 367 | 367 | 100 | 19198 | 23216 | 82.7 |
| **9**: Active Only * | 0.0001 | 0.02 s | 157153 | 157153 | 100 | 367 | 367 | 100 | N/A | N/A | N/A |

**Table 1.** Planner test cases for the case study on LA, TX, CONUS fire dates

The science rewards section shows the total rewards collected in the plan using the objective equation (1), the total available rewards, and percentage of available rewards collected in the plan. Key performance metrics show that 98-100% of all available rewards are collected. 100% of active fire targets and ~80-90% of pre-fire targets are covered. The solver schedules active fire targets immediately (since mandatory), then spends the remaining solve time squeezing in the pre-fire targets to the storage capacity allowed. When the MIP solver is applied to pre-fire targets only using WLFP-based rewards, as a use case not shown in Table 1 [22], it captures >99% of the targets since there are no active fires to compete for storage. Table 1 also confirms that solutions with a relaxed gap tolerance of 1e-2 collected nearly the same science reward % in ~30min as the 1e-4 cases that timed out at 3 hours. As evident from the near maximum rewards collected, the uncaptured targets correspond to very low rewards. Based on cleared buffer, we expect observation-to-downlink latency to be <24h. When added to pre- and post-processing expected in Section 3 and 5 respectively, the end-to-end workflow execution is expected to fit within 30h.

## 5       Data Retrievals and Assimilation

To model plan execution, CYGNSS Mission Operations Center (MOC) commanded rawIF mode over satellite tracks that best matched the geographic polygons the Planner output would have observed, and downlinked the data when GS was available. This step was manual, non-optimal, and caused lengthy delays to collection and downlink. Automation engineering and operationalization are expected to improve execution. Regardless, the AI/ML technologies proposed create novel products like BAMs for active fires, SM for pre-fire conditions, and ML-based fire forecasts and show quantifiable impact to state of art. These improved forecasts support IMTs and may be used toward Section 3.1 to inform responsive observations in the next planning horizon.

### 5.1      Burned Area Maps (BAMs)

BAMs with high temporal, medium spatial resolution and low latency are being developed to be ingested by the wildfire simulation model, per the framework in Fig. 5, from GNSS-R data using the CYGNSS DDMs and an ML model called XGBoost [13]. The training features for the XGBoost binary classification model are extracted from the nominal mode data, consisting of 17×11 delay-Doppler bins. The ancillary data layers are collected from Soil Moisture Active Passive (SMAP) [23] and Shuttle Radar Topography Mission (SRTM) [24]. In addition to the features considered in [25], the novel addition of second-order texture information computed from DDMs as training features is hypothesized (and proven) to improve the XGBoost model. The texture analysis includes calculations of dissimilarity, entropy, and correlation of the gray-level co-occurrence matrix [26] formed from DDMs. The outputs of the ML model are the burn area binary classification, and a confidence level associated with the prediction. The model trains on 200k coordinates in <10 min, predicts 100 coordinates in <1min and uses memory ~1 MiB.

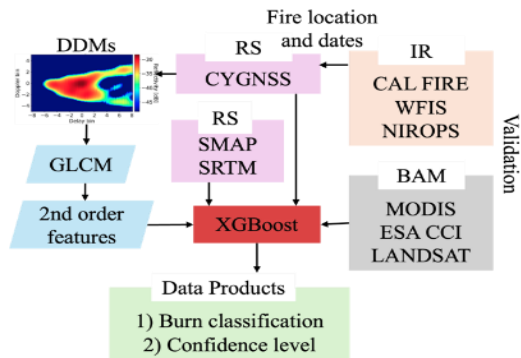

**Fig. 5.** Overall framework for burned area mapping. RS: Remotely sensed data; IR: Infrared data; GLCM: Grey-level co-occurrence matrix.

The ML model is validated using the set-theoretic union of three different burned area products found in the literature [27-29]. We also validated our ML approach in three different study areas, including California and Texas wildfires from CONUS and a region in northern Angola prone to periodic agricultural burns. See Fig. 6 for examples. For our case study, rawIF mode data were available for the 2024 TX and 2025 LA fires in addition to the nominal mode [30]. With the higher resolution of raw IF DDMs (111×69 delay-Doppler bins), we saw a 24% improvement in results from texture analysis when compared to results with just nominal mode data [31]. We saw a 15% boost in the recall of ML model classifications when rawIF was included in training along with nominal mode data. Execution of the optimal plans in a timely manner is expected to improve the collection of raw IF data and BAM classifications even further.

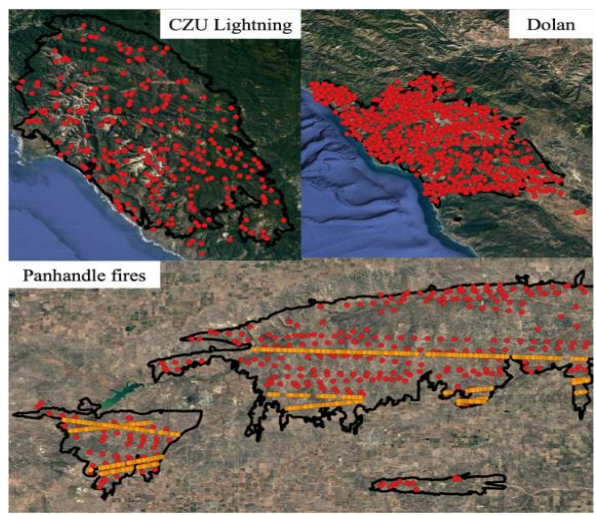

**Fig. 6.** Examples of BAM from XGBoost predictions for CA and TX fires. Red dots are nominal predictions. Yellow dots are raw IF mode predictions.

## 5.2     Active Fire Predictions (WRFx Simulation)

Fire Arrival Time (FAT) can be used as an encapsulation of the history and state of a wildland fire simulation in the WRF-SFIRE model [32] to spin up a coupled atmosphere-fire model gradually when initialized from a developed fire or modified by data assimilation [33]. FAT has been historically retrieved using active fire detections from MODIS and VIIRS, NASA's traditional single satellites with visible or near-infrared sensors at latencies less than 6 hours, by separating burning and not burning pixels in time-space using a Support Vector Machine (SVM) [34].

The availability of BAMs from CYGNSS allows the addition of complementary fire detection data to a novel, ML-based fusion framework to improve state-of-the-art estimates. The Neural Network (NN) architecture in Fig. 7 employs a multi-head design with two branches of output: a classification head for fire detection and a regression head for FAT prediction. The model's backbone consists of densely connected layers with batch normalization and dropout layers for regularization. The fusion approach concatenates features from both data types with source identifiers at the input level, allowing the network to learn joint representations from these complementary fire detection methods. The network processes normalized input features and implements a custom loss function that combines binary cross-entropy for fire classification and weighted consistency-based penalty for FAT prediction. The FAT loss component penalizes predictions where the predicted arrival time is inconsistent with an observation. For fire-detected points, the loss penalizes predictions where fire is estimated to have arrived later than observed and penalizes early predictions in no-fire areas. There is no conventional label for the FAT. The overall loss is the weighted sum of fire and no-fire penalties.

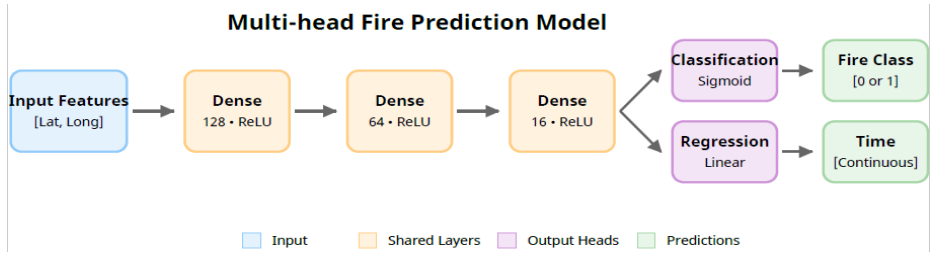

**Fig. 7.** Neural Network architecture to estimate Fire Arrival Time

The training pipeline implements under-sampling to address the imbalance of a small number of fire or burned points vs. a very large number of no-fire or not burned points, along with confidence-based filtering (threshold>70) to enhance data quality. The framework is trained through a custom training loop that supports early stopping with a patience of 50 epochs, utilizing the Adam optimizer with a learning rate of 1e-4. The data fusion and FAT estimates from the NN are seen in Fig. 8 the 2024 Smokehouse Creek fire. MODIS/VIIRS data shows the active fire concentrated during the first two days of the fire, complemented by CYGNSS burned area detections starting from day two. The higher versatility of our NN, particularly with Rectified Linear Unit (ReLU) activation, is expected to enable a more accurate representation of FAT in the presence of sharp changes, such as fire suppression, compared to the SVM solution [34], which is fundamentally limited by the properties of the kernel. These improved forecasts will be integrated into data assimilation cycles within WRFx, as described in Ref [35].

CYGNSS microwave data, which can measure through smoke and clouds (unlike MODIS/VIIRS), add further value to the model output. Initial results from the NN classification of burnt area (Table 2) show that training on CYGNSS improves estimates from VIIRS/MODIS Active Fires (AF) alone, and adding RawIF data improves it further. For FAT estimation, the addition of CYGNSS data did not significantly improve results, because burned area detections appear more than a day after the fire starts. If the MILP planner's schedule is executed in a timely way as proposed in our ConOps instead of approximated by the MOC with resource-constrained delays, we expect the collected CYGNSS data to further improve predictions. Future metrics include Fire

Arrival Time Prediction Accuracy as a function of clouds and reduction in model forecast error when using CYGNSS data in WRFx.

| Training Set | Test Set | Accuracy |
|---|---|---|
| AF | AF | 72.08% |
| AF+CYGNSS Nominal | AF | 84.5% |
| AF+CYGNSS RawIF | AF | 86.04% |

**Table 2.** Table 2. NN performance for the 2024 TX fire. Input features were geographic coordinates, output label was burned at any time during the fire episode. Accuracy is the fraction of test locations classified correctly as Fire/No Fire.

The model demonstrates exceptional efficiency with only ~10k trainable parameters and model size of 0.08 MB, allowing it to achieve impressive inference speeds of 0.17 ms/sample on a single core of EPYC 7502 while maintaining low latency percentiles across various hardware platforms. The minimal resource requirements (3.39 MB memory usage, 2.90 W power consumption) make this model also suitable for low-SWaP deployments on edge devices.

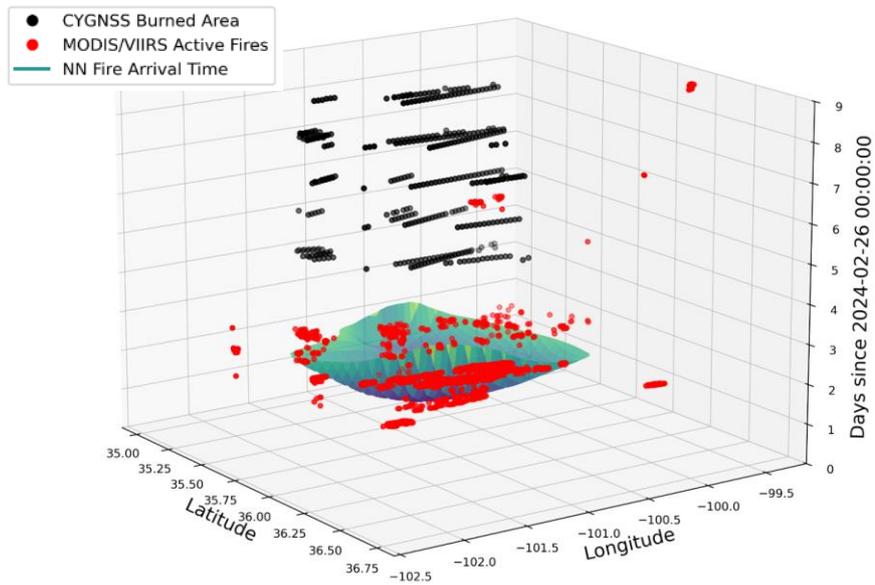

**Fig. 8.** Spatio-temporal evolution inferred for the 2024 TX Panhandle Smokehouse Creek Fire. CYGNSS burned area detections on specular tracks come in later than the MODIS and VIRRS Active Fires detections. Fire Arrival Time contour, estimated by the proposed NN with a bias correction applied to all data is shown. No-fire detections on CYGNSS specular point tracks and in MODIS/VIIRS granules are not shown for visibility.

### 5.3     Pre-Fire Predictions (USGS Fire Danger)

Multiple algorithms have been developed to estimate surface SM from CYGNSS observations [30,36]. The nominal mode produces a daily SM product with 9 km and 36 km resolution, estimated using linear regression [37], and delivers it publicly [38]. SM products from rawIF data use ML methods [39], specifically NGBoost/XGBoost, and are experimentally available in our workflow. Previous research has indicated a significant correlation between SM and fuel moisture which affects the estimation of wildland fire potential [40], however neither have been incorporated into USGS' pre-fire danger predictions.  We have developed an enhanced WFPI product (WFPI-x) by integrating SM estimates from CYGNSS into WFPI and subsequently derived WLFP-x and WFSP-x from WFPI-x (defined in section 3.2). Details on all models are described in [41], and validation shows expected trends in all *-x products. Since rawIF mode improves the accuracy of retrievals [39], execution of the optimal planner schedules for rawIF observations is expected to improve *-x products, leading to improved priorities set by Section 3.1, thereby improving our understanding of pre-fire conditions with every cycle.

## 6      Discussion and Future Work

We demonstrated an AI-driven spaceborne ConOps that allows the intelligent and responsive collection of novel sensor data that can be processed, and used to track, and forecast the spread of active wildfires and pre-fire danger. Novel AI elements are used in many parts of the concept – ML-based fire predictors, MILP-based planner, ML-based post-processed data products, NN-based assimilations and improved predictions. In addition to enabling wildfire monitoring with novel data and improved performance due to responsive feedback loop, these AI elements demonstrate benefit over current state of art, individually and together. Many sustainability considerations, such as low SWaP for sensors and inference, energy-efficient AI deployment, efficient onboard planning, and low latency of feedback, have been included in the system architecture and ConOps. We also found that, based on runtimes and computational requirements, the concept and its components are feasible to scale with heavier wildfire loads.

   While we demonstrate the concept and models on CONUS fires, it is generalizable geospatially with differing considerations for various components. The spacecraft simulator (EO-Sim) and planner models are global. SM and BAM products use global spacecraft sensor and ancillary data, therefore applicable worldwide (as evident in validation outcomes in Angola, Africa). Active fire simulations for initial forecasts depend on the quality and resolution of fire weather and fuels data, e.g. HRRR and RAWS are local to CONUS,  but alternative sources are available in other parts of the world. Active fire assimilation needs appropriate sensor data (CYGNSS, MODIS, VIIRS products are global) and WRFx, which can be used internationally if  reliable and high-resolution fire behavior fuel model map is available for the region of interest. A global MODIS-based "landuse" product has been used as an alternative in fuel behavior models with a cross-walk of classes. WRF-SFIRE has been demonstrated internationally, e.g. in Greece [42], India [43], Chile, Canada. For pre-fire danger predictions, WFPI, and therefore all dependencies and enhancements described, is a CONUS product. However, similar products are feasible given the availability of fuels and weather data as with active fire cases.

Future work includes ML model improvements for both pre/active fire danger predictors, adding uncertainty quantification to the predictions, and post-assimilation FAT forecasts and using them toward improvement of the AI planner objectives and constraints. We plan to use more GNSS-R sensors beyond CYGNSS, e.g. commercial providers such as Spire Global and Muon Space, which may provide more onboard resources and further improve model outcomes. Engineering improvements to workflow automation will also ensure maximal benefit of the ConOps execution and products.

## Ethics Statement and Acknowledgments

There are no ethical issues regarding the research being reported and the broader ethical impact of the work.

All the code and data described will be updated on https://github.com/dshield-proj.

This manuscript is based on work supported by the National Aeronautics and Space Administration (NASA) using primary grant 80NSSC23K1118, and partially on 80NSSC23K1344, 80NSSC22K1717, and 80NSSC22K1405.

## References

1. Doerr, Stefan H., and Cristina Santín. "Global trends in wildfire and its impacts: perceptions versus realities in a changing world" Philosophical Transactions of the Royal Society B: Biological Sciences 371.1696 (2016): 20150345.
2. V.C. Radeloff, D.P. Helmers, H.A. Kramer, M.H. Mockrin, P.M. Alexandre, A. Bar-Massada, V. Butsic, T.J. Hawbaker, S. Martinuzzi, A.D. Syphard, & S.I. Stewart, Rapid growth of the US wildland-urban interface raises wildfire risk, Proc. Natl. Acad. Sci. U.S.A. 115 (13) 3314-3319.
3. National Interagency Fire Center Statistics website https://www.nifc.gov/fire-information/statistics and open data website https://data-nifc.opendata.arcgis.com/pages/wfigs-page; Retrieved 9 February 2025
4. Nicholas Bogel-Burroughs, Jacey Fortin, and Anna Betts. Texas Fires Span Over 1.2 Million Acres. Here Is What We Know. In The New York Times, 7 Mar. 2024. Website: https://www.nytimes.com/article/texas-smokehouse-creek-fire.html. Accessed 31 Jan. 2025
5. Soumya Karlamangla, Jesus Jiménez, Yan Zhuang, Kate Selig and Rachel Nostrant. What We Know About the Wildfires in Southern California. In The New York Times, 8 Jan. 2025, https://www.nytimes.com/2025/01/08/us/wildfires-los-angeles-california.html Accessed 31 Jan. 2025.

6. X. Wang, G. Wu, L. Xing and W. Pedrycz, "Agile Earth Observation Satellite Scheduling Over 20 Years: Formulations, Methods, and Future Directions," in IEEE Systems Journal, vol. 15, no. 3, pp. 3881-3892, Sept. 2021.

7. Christopher S. Ruf, Clara Chew, Timothy Lang, Mary G. Morris, Kyle Nave, Aaron Ridley, and Rajeswari Balasubramaniam. "A new paradigm in earth environmental monitoring with the CYGNSS small satellite constellation." Scientific reports 8, no. 1 (2018): 8782.

8. J. Mandel, M. Vejmelka, A. Kochanski, A. Farguell, J. Haley, D. Mallia, and K. Hilburn. An Interactive Data-Driven HPC System for Forecasting Weather, Wildland Fire, and Smoke. 2019IEEE/ACM HPC for Urgent Decision Making (UrgentHPC), Denver, CO, USA, 2019, pages 35-44.

9. Nag, S., Ravindra, V., Levinson, R., Moghaddam, M., Nelson, K., Mandel, J., Kochanski, A., Caus, A.F., Melebari, A., Kannan, A. and Ketzner, R., 2024, July. Distributed Spacecraft with Heuristic Intelligence to Monitor Wildfire Spread for Responsive Control. IEEE International Geoscience and Remote Sensing Symposium 2024 (pp. 699-703).

10. S. A. Chien, A. G. Davies, J. Doubleday, D. Q. Tran, D. Mclaren, W. Chi, A. Maillard, "Automated Volcano Monitoring Using Multiple Space and Ground Sensors," Journal of Aerospace Information Systems, vol. 17, no. 4, pp. 214-228, 4 2020.

11. S. Roy-Singh, A.P. Li, V. Ravindra, R. Lammers, "Agile, Autonomous Spacecraft Constellations with Disruption Tolerant Networking to Monitor Precipitation and Urban Floods", Robotics Science and Systems – Space Robotics Workshop, June 2025.

12. WRFx simulation database portal, San Jose State University's Wildfire Interdisciplinary Center. Retrieved Feb 2025 https://wrfx.org/ or https://demo.openwfm.com/sj/

13. Tianqi Chen, and Carlos Guestrin. "Xgboost: A scalable tree boosting system." In Proceedings of the 22nd ACM SIGKDD International Conference on Knowledge Discovery and Data Mining, pp. 785-794. 2016.

14. USGS Fire Danger Forecast. https://www.usgs.gov/fire-danger-forecast Retrieved 7 Feb 2025.

15. Robert Burgan, Robert Klaver, Jacqueline Klaver. Fuel Models and Fire Potential from Satellite and Surface Observations. International Journal of Wildland Fire 8 (3): 159-170.

16. Haiganoush Preisler, David Brillinger, Robert Burgan, J.W. Benoit. Probability based models for estimation of wildfire risk. International Journal of Wildland Fire 13: 133-142.

17. Haiganoush Preisler, Robert Burgan, Jeffery Eidenshink, Jacqueline Klaver, Robert Klaver. Forecasting distributions of large federal-lands fires utilizing satellite and gridded weather information. International Journal of Wildland Fire 18: 508-516.

18. Vinay Ravindra, Ryan Ketzner, and Sreeja Nag. Earth observation simulator (EO-Sim): An open-source software for observation systems design. In 2021 IEEE International Geoscience and Remote Sensing Symposium IGARSS, pp. 7682-7685. IEEE, 2021.

19. Earth Observation Simulator (EO-Sim) https://github.com/EarthObservationSimulator . Retrieved 31 Jan. 2025.

20. Levinson, R., Niemoeller, S., Nag, S., & Ravindra, V., Planning satellite swarm measurements for earth science models: comparing constraint processing and MILP method, Proceedings of the International Conference on Automated Planning and Scheduling, ICAPS 2022, Vol. 32, pp. 471-479.

21. Gurobi Optimization and Decision-making Technology. https://www.gurobi.com/ . Accessed online 8-Jan-2025.

22. Levinson, R., Ravindra, V., Roy-Singh, S., Optimal Planning to Coordinate Science Data Collection and Downlink for a Constellation of Agile Satellites with Limited Storage, Proceedings of the International Joint Conference on Artificial Intelligence, IJCAI 2025.

23. Dara Entekhabi, Eni G. Njoku, Peggy E. O'neill, Kent H. Kellogg, Wade T. Crow, Wendy N. Edelstein, Jared K. Entin et al. "The soil moisture active passive (SMAP) mission." Proceedings of the IEEE 98, no. 5 (2010): 704-716.

24. Earth Resources Observation and Science (EROS) Center, "Shuttle Radar Topography Mission 1 Arc-Second Global," 2018.

25. A. Kannan, A. Melebari, G. Tsagkatakis, K. Nelson, V. Ravindra, S. Nag, and M. Moghaddam. "Mapping Wildfire Burned Area Using GNSS-Reflectometry in Densely Vegetated Regions with Complex Topography: A Machine Learning Approach." IEEE International Geoscience and Remote Sensing Symposium 2024.

26. R. M. Haralick, K. Shanmugam, and Its' Hak Dinstein. "Textural features for image classification." IEEE Transactions on systems, man, and cybernetics 6 (1973): 610-621.

27. Louis Giglio, Luigi Boschetti, David P. Roy, Michael L. Humber, and Christopher O. Justice. "The Collection 6 MODIS burned area mapping algorithm and product." Remote sensing of environment 217 (2018): 72-85

28. E. M. Chuvieco, L. Pettinari, J. Lizundia-Loiola, T. Storm, and M. Padilla Parellada. "ESA fire climate change initiative (Fire_cci): MODIS Fire_cci burned area pixel product, version 5.1.", (2018)

29. Todd J. Hawbaker, Melanie K. Vanderhoof, Gail L. Schmidt, Yen-Ju Beal, Joshua J. Picotte, Joshua D. Takacs, Jeff T. Falgout, and John L. Dwyer. "The Landsat Burned Area algorithm and products for the conterminous United States." Remote Sensing of Environment 244 (2020): 111801

30. Hugo Carreno-Luengo, Christopher S. Ruf, Scott Gleason, and Anthony Russel. "A New Multi-Resolution CYGNSS Data Product for Fully and Partially Coherent Scattering." IEEE Transactions on Geoscience and Remote Sensing (2023).

31. A. Kannan, A. Melebari, G. Tsagkatakis, A. F. Caus, K. Nelson, A. Kochanski, V. Ravindra, S. Nag, C. Ruf, M. Moghaddam "Enhancements to Mapping Wildfire Burned Areas using GNSS-Reflectometry Raw Intermediate Frequency data", IEEE International Geoscience and Remote Sensing Symposium 2025.

32. J. Mandel, S. Amram, J. D. Beezley, G. Kelman, A. K. Kochanski, V. Y. Kondratenko, B. H. Lynn, B. Regev, and M. Vejmelka. "Recent advances and applications of WRF-SFIRE". Natural Hazards and Earth System Science, 14(10):2829–2845, 2014.

33. J. Mandel, A. K. Kochanski, M. Vejmelka, J. D. Beezley. "Data assimilation of satellite fire detection in coupled atmosphere-fire simulations by WRF-SFIRE." in Advances in Forest Fire Research, pages 716–724. Coimbra University Press, 2014.

34. Angel Farguell, Jan Mandel, James Haley, Derek V. Mallia, Adam Kochanski, and Kyle Hilburn. "Machine learning estimation of fire arrival time from level-2 Active Fires satellite data." Remote Sensing, 13(11):2203, Jun 2021.

35. A. K. Kochanski, K. Clough. A. Farguell. D. V. Mallia, J. Mandel, K. Hilburn, "Analysis of methods for assimilating fire perimeters into a coupled fire-atmosphere model", Frontiers in Forests and Global Change, 6, 1203578, 2023.

36. Wu, Xuerui, Wenxiao Ma, Junming Xia, Weihua Bai, Shuanggen Jin, and Andrés Calabia. 2020. "Spaceborne GNSS-R Soil Moisture Retrieval: Status, Development Opportunities, and Challenges." Remote Sensing 13 (1): 45.

37. Chew, Clara, and Eric Small. 2020. "Description of the UCAR/CU Soil Moisture Product." Re-mote Sensing 12 (10): 1558.

38. CYGNSS. 2024. "CYGNSS Level 3 Soil Moisture Version 3.2." NASA Physical Oceanography Distributed Active Archive Center. https://doi.org/10.5067/CYGNU-L3S32

39. G. Tsagkatakis, A. Melebari, R. Akbar, J. D. Campbell, E. Hodges, and M. Moghaddam, "Uncertainty Quantification in Machine Learning Based Retrieval of Soil Moisture from

GNSS-R Observations", IEEE International Geoscience and Remote Sensing Symposium, Athens, Greece, 2024

40. E. Krueger, M. Levi, K. Achieng, J. Bolten, J. Carlson, N. Coops, et al., "Using soil moisture information to better understand and predict wildfire danger: a review of recent developments and outstanding questions," International Journal of Wildland Fire, vol. 32, pp. 111-132, 2023.

41. K. Nelson, S. Roy-Singh, V. Ravindra, M. Moghaddam, A. Kannan, A. Melebari, "Integration of GNSS-R derived Soil Moisture into the USGS Wildland Fire Potential Index", IEEE International Geoscience and Remote Sensing Symposium 2025.

42. Kartsios, S., Karacostas, T., Pytharoulis, I., & Dimitrakopoulos, A. P. (2021). Numerical investigation of atmosphere-fire interactions during high-impact wildland fire events in Greece. Atmospheric Research, 247, 105253.

43. Kale, M. P., Meher, S. S., Chavan, M., Kumar, V., Sultan, M. A., Dongre, P., ... & Roy, P. S. (2024). Operational Forest-Fire Spread Forecasting Using the WRF-SFIRE Model. Remote Sensing, 16(13), 2480.